# Numerical Words and Linguistic Loops: The Perpetual Four-Letter Routine

Krishna C. Polavaram

Department of Civil and Environmental Engineering, University of Illinois at Urbana-Champaign, 205 N. Mathews, Urbana, IL 61801.

## Abstract

This study presents a fascinating linguistic property related to the number of letters in words and their corresponding numerical values. By selecting any arbitrary word, counting its constituent letters, and subsequently spelling out the resulting count and tallying the letters anew, an unanticipated pattern is observed. Remarkably, this iterative sequence, conducted on a dataset of 100,000 random words, invariably converges to the numeral four (4), termed the "Linguistic Loop (LL) constant". Examining 73 languages utilizing the Latin alphabet, this research reveals distinctive patterns. Among them, 28 languages exhibit LL-positive behavior adhering to the established property, while 31 languages deviate as LL-negative. Additionally, 13 languages display nuanced tendencies: eight feature two LL constants (bi-positivity), and five feature three constants (tri-positivity). This discovery highlights a linguistic quirk within Latin alphabet-based language number-word representations, uncovering an intriguing facet across diverse alphabetic systems. It also raises questions about the underlying linguistic and cognitive mechanisms responsible for this phenomenon.

## Introduction

The intricate interplay between language and mathematics has intrigued scholars for centuries, yielding insights into the cognitive underpinnings of human communication and thought. As the linguistic and numerical domains intersect, fascinating avenues for exploration emerge, inviting us to probe the depths of linguistic patterns and their potential mathematical basis.

Linguistic patterns are integral to the fabric of human expression, reflective of cognitive processes that shape language acquisition, comprehension, and production. The study of linguistic patterns and their potential connections to mathematical constructs has sparked a rich discourse within cognitive linguistics and psycholinguistics. Insights from studies such as Lakoff & Núñez (2000) on conceptual metaphors in mathematics, as well as work by Gordon (2004) on the relationship between linguistic determinism and numerical cognition, highlight the intricate interplay between linguistic structures and mathematical thought. Moreover, research by Dehaene & Cohen (1997) underscores the cross-domain connections between number processing and language, emphasizing the role of linguistic representations in shaping numerical thought. Additionally, Whorfian hypothesis-inspired studies, such as those by Everett (2005) and (Gelman & Gallistel 2004), have unearthed linguistic influences on numerical cognition, shedding light on how language can potentially reframe our numerical experiences. Finally, the studies of linguistic patterns and cognitive frameworks within linguistic anthropology and cognitive linguistics from work by Lakoff & Johnson (1980) on metaphorical mappings and work by Gentner, Imai & Boroditsky (2002) on exploration of time metaphors inform our understanding of how linguistic structures reflect and shape cognitive processes.

Within this context, this study reports a discovery that draws inspiration from these interdisciplinary intersections. It centers on the iterative transformation of words into numbers and back again—converting word lengths to numerical values, spelling out these numbers, and subsequently assessing the letter count of these spellings. This iterative process unveils an astonishing regularity—a consistent convergence to the numeral four, regardless of the initial word chosen. We denote this phenomenon as the Linguistic Loop Constant (LLC). Furthermore, I meticulously analyze 73 languages that employ the Latin alphabet. Among these languages, a captivating linguistic pattern emerges. While 28 languages manifest LL-positive behavior, adhering to the established property, 31 languages display LL-negative tendencies, deviating from the pattern. Further examination uncovers nuanced behaviors—eight languages feature two LLCs, indicative of bi-positivity, while five languages showcase three constants, demonstrating tri-positivity. Navigating through this uncharted territory, this investigation sheds light on the intricate interplay between linguistic structures and numerical concepts. Furthermore, this discovery prompts contemplation on the potential interplay of human cognition and expression, offering a glimpse into the innate symmetry that emerges when language and numbers converge. Finally, this study contributes to the existing discourse at the crossroads of linguistics and mathematics, inviting further exploration into the cognitive architecture that unifies these seemingly distinct realms of human understanding.

## Methodology

This section delineates the comprehensive methodology used to explore the linguistic phenomenon related to the number of letters in words and their numerical values, leading to the identification of the "Linguistic Loop constant." The study employs a multi-faceted approach encompassing quantitative analysis, computational routines, and corpus-based investigations across diverse languages employing the Latin alphabet.

### Corpus Compilation and Data Sources

The data for this study is derived from two distinct sources, each serving a crucial role in uncovering the linguistic patterns. Firstly, the MIT Wordlist Tool was used to establish a foundational dataset for analysis by generating a collection of 100,000 English words. This wordlist encompasses a wide spectrum of vocabulary, facilitating an examination of linguistic relationships. Secondly, to expand the analysis across different languages, the study incorporated translations of the wordlist. The "googletrans" and "requests" Python packages were employed to automatically translate the English words into the respective languages, creating new datasets of words for further examination. In addition, Google Translate was used to obtain the word spellings for numbers ranging from one to the length of the longest word in each chosen language. These translated numerical representations constitute a unique linguistic corpus, augmenting the analysis beyond English.

**Quantitative Analysis and Computational Routine**

The MIT wordlist-derived English word dataset laid the groundwork for analysis. Each word's letter count was meticulously quantified, enabling the construction of a probability density distribution characterizing word lengths. Further, an iterative computational routine was executed using a custom Python code which was developed for iterative transformations of words into numbers and back again. This process involved translating word lengths to numerical values, spelling out these numbers, and subsequently assessing the letter count of the resulting spellings. The proportion of words with a length of 4 was tracked after each iteration, alongside the proportions of words of varying sizes. Next, the results of the iterative routine were visually represented through graphs, allowing for the identification of patterns and trends. Thorough analysis of these visualizations aided in recognizing the emergence of the "Linguistic Loop constant."

**Corpus Analysis Across Multiple Languages**

A diverse set of 73 languages utilizing the Latin alphabet were chosen for exploration. These languages encompass a wide array of linguistic and cultural backgrounds. Next, Google Translate was harnessed to obtain the word spellings for numbers in each selected language. This collection formed the basis of the expanded computational routines. Further, the established iterative computational routine, initially developed for English, was systematically applied to the translated numbers in each language. This facilitated the identification of linguistic patterns analogous to those observed in English. Finally, examination of the iterative routine results for each language led to the identification of distinct linguistic behaviors—LL-positive, LL-negative, bi-positive, and tri-positive. These categorizations were critical in highlighting diverse linguistic responses to the iterative process. To illustrate the behavior of the different linguistic phenomenon, translations were generated for languages including Swedish (LL bi-positive), Norwegian (LL tri-positive), and Indonesian (LL negative).

**Software Tools and Validation**

Python programming language, supplemented by relevant libraries for data manipulation, visualization, and computation, underpinned the computational analyses. Independent validation procedures were implemented to ensure the precision of the computational routines, bolstering the reliability of the outcomes. The multifaceted methodology employed in this study encompasses data compilation from established linguistic sources, custom computational routines, and systematic exploration of diverse linguistic contexts. By combining quantitative analysis and corpus-based investigations, this methodology facilitated a robust investigation to unravel the LLC while shedding light on the intricate interplay between linguistic structures and numerical representations.

## Analysis and Discussion

This section thoroughly examines the findings obtained from the quantitative analysis and computational routines, shedding light on the behavior of the LLC across different languages and its implications for linguistic patterns and numerical representations. The initial analysis of the English wordlist dataset revealed intriguing patterns and trends related to the LLC. Fig. 1a presents the probability density distribution of the number of letters in English words. The distribution showcased a peak at a word length of 7, resembling a normal distribution. Fig. 1b and Fig. 1c illustrated the proportion of words with a length of 4 and the proportion of words of all sizes, respectively, after each iteration. The histogram's bars denote word lengths in different segments. A significant observation was the consistent increase in the proportion of words with a length of 4 as the number of iterations grew, while the proportion of words with other lengths decreased. This behavior indicated the convergence phenomenon toward the LLC.

**Table 1.** Classification of languages using the Latin alphabet based on LL-positivity and LLCs: Languages using the Latin alphabet with LLCs ranging from 2-6 as LL-positive, and those that are LL-negative.

| LL Constant | Languages using Latin alphabet | Count |
|:---:|:---|:---:|
| 2 | Kurdish | 1 |
| 3 | Afrikaans, Czech, Ewe, Igbo, Irish, Italian, Slovak, Scottish Gaelic, Slovenian | 9 |
| 4 | Aymara, Dutch, English, Estonian, Filipino, German, Hungarian, Mizo, Oromo, Sundanese | 10 |
| 5 | Bambara, Finnish, Galician, Hawaiian, Romanian, Yoruba | 6 |
| 6 | Corsican, Lingala | 2 |
| LL-negative | Albanian, Azerbaijani, Basque, Haitian creole, Guarani, Icelandic, Lloko, Indonesian, Karindwi, Latin, Luxembourgish, Malay, Maori, Northern Sotho, Nyanja, Polish, Portuguese, Quechua, Samoan, Shona, Southern Sotho, Spanish, Swahili, Tsonga, Turkish, Turkmen, Uzbek, Vietnamese, Western Frisian, Xhosa, Zulu | 31 |

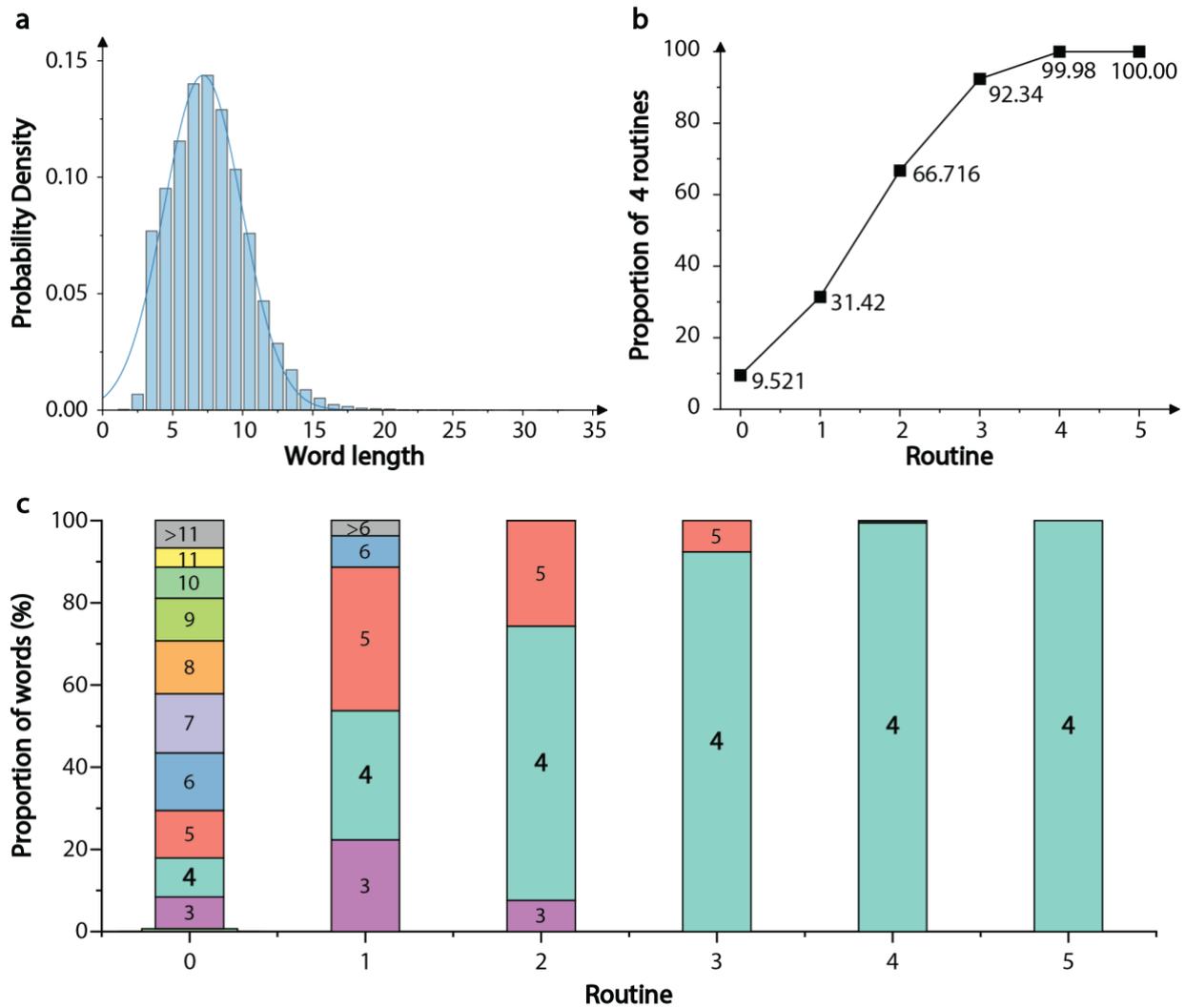

**Fig. 1**. a) Probability density distribution of the number of letters in the 100,000 words generated. b) Proportion of words with a length of 4 obtained after performing each routine. c) Proportion of words of all lengths obtained after performing each routine. The word length is indicated on the bars in every segment. Color schemes were adapted from ColorBrewer to have an unbiased representation (Brewer, Hatchard & Harrower 2003).

**Linguistic Diversity in Languages**

The study's exploration extended beyond English to examine the LL phenomenon across a spectrum of languages employing the Latin alphabet. This analysis led to the classification of languages based on their LL behaviors: LL-positive and LL-negative languages. Table 1 outlines the classification of languages based on their LL-positivity. Languages that exhibit LL-positive behavior, with LLCs ranging from 2 to 6, adhere to the established pattern. These languages demonstrate a consistent convergence towards the LL constant, reflecting a linguistic symmetry.

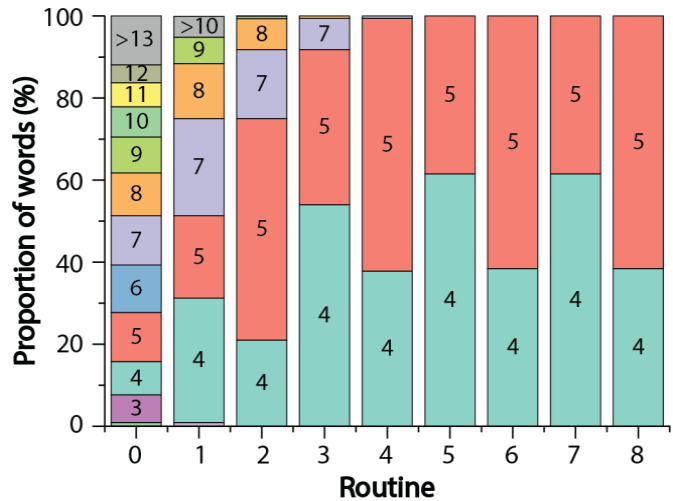

**Fig. 2** Proportion of words of all lengths obtained after performing each routine for Indonesian, a LL-negative language. Notably, there is no convergence in word lengths.

On the other hand, the analysis also identified LL-negative languages, which deviate from the convergence pattern observed in LL-positive languages. These languages display distinct behaviors, presenting intriguing and inconsistent departures from the established phenomenon. Fig. 2 illustrates the results from various routines for Indonesian, an LL-negative language. Notably, there is no convergence in the proportion of words of different lengths, even after 8 routines. This illustrates the nature of a LL-negative language.

**Bi-Positive and Tri-Positive Languages**

Intriguing linguistic behaviors emerged within specific language groups, revealing the intricacies of linguistic representations. Table 2 provides a list of languages displaying bi-positivity, featuring two PCs and tri-positivity, featuring three PCs. This phenomenon adds a layer of complexity to the linguistic patterns observed, underscoring the varied ways in which languages represent numerical values. Further, the tri-positivity behavior, highlighted in Table 2, introduces an additional dimension to the linguistic analysis. Fig. 3a illustrates the convergence phenomena for Swedish, a bi-positive language characterized by PCs of 3 and 4. Likewise, Fig. 3b illustrates the convergence phenomena for Norwegian, a tri-positive language characterized by PCs of 2, 3, and 4. This nuanced analysis unveils the intricate relationships between linguistic diversity and numerical regularities within diverse linguistic frameworks.

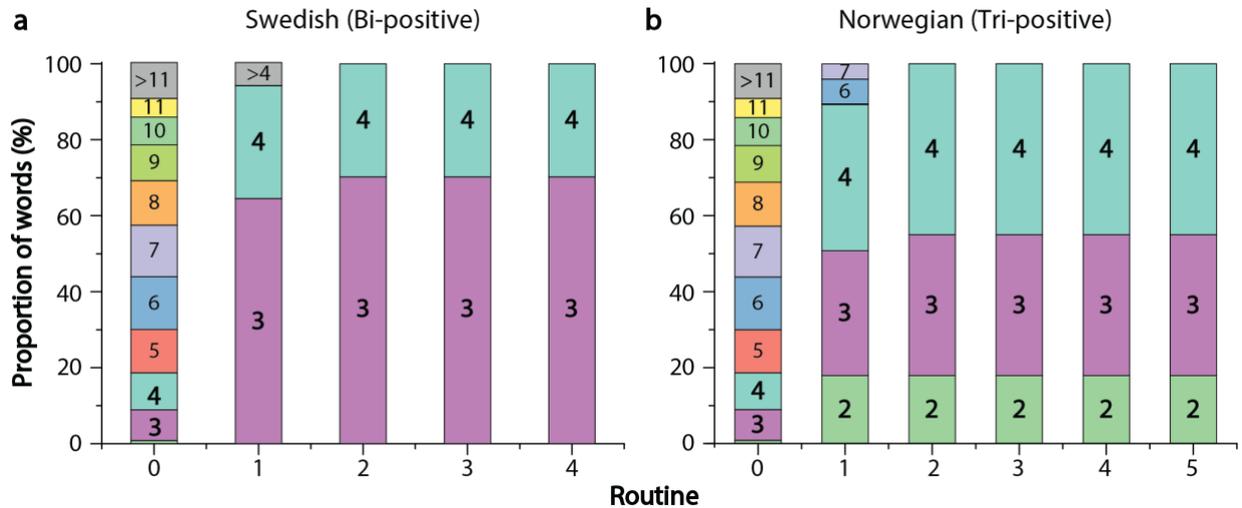

**Fig. 3** Illustration of the convergence phenomena for Swedish and Norwegian, representing bi-positive and tri-positive languages, respectively. Notably, a proportion of words of various lengths exist initially (routine 0), but with increasing routines these proportions segregate into 2 or 3 distinct word lengths, depending on the language type.

**Table 2.** List of LL bi-positive and tri-positive languages using the Latin alphabet and the corresponding LLCs.

| Type | Languages using Latin alphabet | LLC 1 | LLC2 | LLC3 |
|---|---|---|---|---|
| Bi-positive | Aymara | 4 | 11 | - |
| | Danish | 2 | 4 | - |
| | Homng | 2 | 3 | - |
| | Latvian | 5 | 7 | - |
| | Swedish | 3 | 4 | - |
| | Maltese | 4 | 5 | - |
| | Somali | 4 | 7 | - |
| | Welsh | 3 | 6 | - |
| Tri-positive | Norwegian | 2 | 3 | 4 |
| | Esperanto | 2 | 3 | 4 |
| | Gand | 4 | 5 | 7 |
| | Hausa | 3 | 4 | 5 |
| | Lithuanian | 2 | 6 | 7 |

## Conclusion

The analysis of the linguistic phenomenon surrounding the "Linguistic Loop constant" engenders thoughtful reflection on the relationship between linguistic and numerical representations. The convergence patterns observed in both English and various languages emphasize the intricate connection between linguistic structures and numerical concepts. The distinct linguistic behaviors across languages underscore the diversity and complexity of linguistic representations, highlighting the role of linguistic and cognitive mechanisms in shaping numerical thought. The analysis further prompts consideration of the underlying mechanisms that drive the convergence and variation in the "Linguistic Loop" phenomenon across languages. This phenomenon draws attention to the intertwined nature of language and mathematics, offering a glimpse into the interwoven fabric of human cognition and expression. The study's findings contribute to the evolving discourse at the crossroads of linguistics and mathematics, inviting further exploration into the cognitive architecture that unifies these seemingly distinct realms of human understanding.

## References

Brewer, Cynthia A., Geoffrey W. Hatchard & Mark A. Harrower. 2003. ColorBrewer in print: A catalog of color schemes for maps. *Cartography and Geographic Information Science* 30(1). 5–32. https://doi.org/10.1559/152304003100010929.

Dehaene, Stanislas & Laurent Cohen. 1997. Cerebral pathways for calculation: Double dissociation between rote verbal and quantitative knowledge of arithmetic. *Cortex*. Elsevier 33(2). 219–250.

Everett, DanielL. 2005. Cultural constraints on grammar and cognition in Pirahã: Another look at the design features of human language. *Current anthropology*. The University of Chicago Press 46(4). 621–646.

Gelman, Rochel & Charles Randy Gallistel. 2004. Language and the origin of numerical concepts. *Science*. American Association for the Advancement of Science 306(5695). 441–443.

Gentner, Dedre, Mutsumi Imai & Lera Boroditsky. 2002. As time goes by: Evidence for two systems in processing space→ time metaphors. *Language and cognitive processes*. Taylor & Francis 17(5). 537–565.

Gordon, Peter. 2004. Numerical cognition without words: Evidence from Amazonia. *Science*. American Association for the Advancement of Science 306(5695). 496–499.

Lakoff, George & Mark Johnson. 1980. The metaphorical structure of the human conceptual system. *Cognitive science*. No longer published by Elsevier 4(2). 195–208.

Lakoff, George & Rafael Núñez. 2000. *Where mathematics comes from*. Vol. 6. New York: Basic Books.

## Funding details

The author declares there are no funding sources.

## Disclosure statement

The author reports there are no competing interests to declare.